\pdfoutput=1

\documentclass[11pt]{article}

\usepackage[final]{acl}

\usepackage{times}
\usepackage{latexsym}

\usepackage[T1]{fontenc}

\usepackage[utf8]{inputenc}

\usepackage{microtype}

\usepackage{inconsolata}

\usepackage{graphicx}
\usepackage{booktabs}
\usepackage{enumitem}
\usepackage{multirow}
\usepackage[greek,english]{babel}
\usepackage{makecell}

\title{Leveraging LLMs for Translating and Classifying Mental Health Data}

\author{\textbf{Konstantinos Skianis}$^*$, \textbf{A. Seza Doğruöz}$^\#$, \textbf{John Pavlopoulos}$^\dagger$$^\P$\\
$^*$ Department of Computer Science and Engineering, University of Ioannina, Greece\\
$^\#$ LT3, IDLab, Universiteit Gent, Belgium \\
$^\dagger$ Department of Informatics, Athens University of Economics and Business, Greece\\
$^\P$ Archimedes/AthenaRC, Greece\\
\texttt{kskianis@cse.uoi.gr \quad as.dogruoz@ugent.be \quad annis@aueb.gr}
}

\begin{document}
\maketitle
\begin{abstract}
Large language models (LLMs) are increasingly used in medical fields. In mental health support, the early identification of linguistic markers associated with mental health conditions can provide valuable support to mental health professionals, and reduce long waiting times for patients.
Despite the benefits of LLMs for mental health support, there is limited research on their application in mental health systems for languages other than English. 
Our study addresses this gap by focusing on the detection of depression severity in Greek through user-generated posts which are automatically translated from English. Our results show that GPT3.5-turbo is not very successful in identifying the severity of depression in English, and it has a varying performance in Greek as well. 
Our study underscores the necessity for further research, especially in languages with less resources.
Also, careful implementation is necessary to ensure that LLMs are used effectively in mental health platforms, and human supervision remains crucial to avoid misdiagnosis.

\end{abstract}

\section{Introduction}


Mental health issues (e.g.,  depression, anxiety, and post-traumatic stress disorder (PTSD)) are prevalent worldwide and pose significant challenges to public health \cite{who2021}. 
Traditional methods for diagnosing mental health conditions often rely on self-reported surveys, clinical interviews, and standardised assessments conducted by trained professionals \cite{kessler2004}. 
While these methods are effective, they are also resource-intensive, time-consuming, and may not always be accessible to individuals in need, particularly for speakers of languages beyond English. 

In this context, the application of LLMs to detect mental health symptoms from textual data offers a compelling alternative. These models can analyse large volumes of text data (e.g., social media posts, forum discussions, and personal narratives) quickly to identify linguistic markers associated with mental health conditions \cite{guntuku2019, chancellor2019}. 
This capability opens up new avenues for early detection and intervention, providing valuable support to mental health professionals and potentially reaching out to the patients whose symptoms may be overlooked and/or save time (e.g., long waiting times). 


Despite the potential benefits, the performance of LLMs in multilingual mental health symptom detection remains underexplored. Previous studies have primarily focused on English-language datasets, leaving a gap in our understanding of how these models perform in other linguistic contexts \cite{raihan2024mentalhelp}. 
Hence, our work raises the following research questions: 
\begin{itemize}
\setlength\itemsep{0.005cm}
\item Can an LLM accurately predict the severity of mental health conditions from English user-generated posts? 
\item Is the detection performance similar if one automatically translates the English posts to another language (e.g., Greek) with LLMs?
\end {itemize}

\begin{figure*}[t]
  \includegraphics[width=\textwidth]{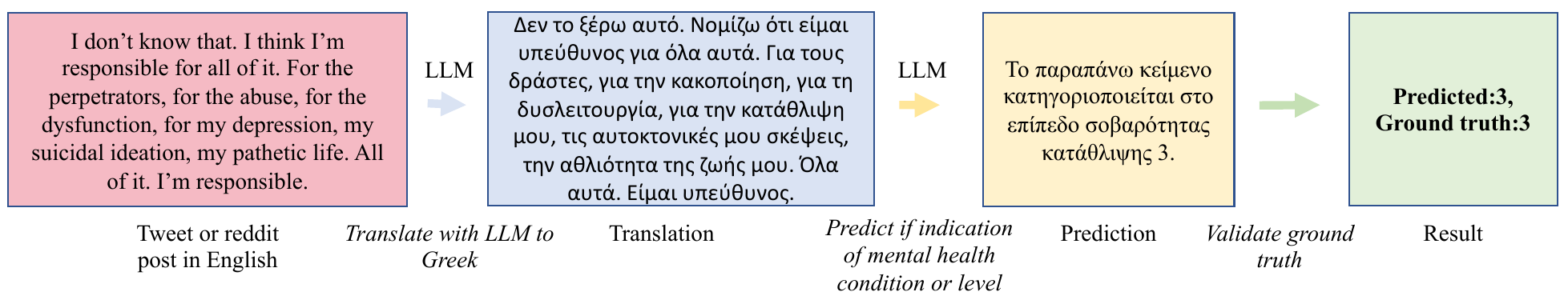} 
  \caption {An illustration of our proposed methodology.}
  \label{proposed}
\end{figure*}

To address these research questions, first, we assess a state-of-the-art multilingual LLM when predicting the severity of mental health in English user-generated posts. Then, we automatically translate these posts from English to Greek, a language for which there are no resources for this task \cite{bakagianni2025systematic},
and re-assess the performance of the LLM.
Our research not only contributes to the development of more robust and inclusive AI-driven mental health diagnostic tools but also emphasises the importance of culturally and linguistically sensitive approaches in mental health care beyond English. The contribution of this work lies into the evaluation of the predictive power of a popular LLM in detecting the severity of depression across English and Greek.



\section{Related work}



LLMs have remarkable accuracy in detecting mental health symptoms by leveraging their ability to understand context and semantics at a deeper level. Examples include BioBERT \cite{lee2020biobert}, and ClinicalBERT \cite{clinicalbert}, which are pre-trained on biomedical corpora or clinical notes. In contrast, models like MentalBERT \cite{ji2022mentalbert}, and DisorBERT \cite{aragon2023disorbert} are pre-trained on mental health-related social media data.
Additionally, research by \citet{benton2017multi} showed that NLP can effectively assess depression and PTSD from clinical notes, further validating the utility of these models in a healthcare setting.

Although the details about training and evaluation are not always transparent, the multilingual capabilities of LLMs enable these models to understand and generate text in various languages. A recent example is the XLM-R model, which has been trained on a vast amount of multilingual data and shows strong performance across multiple languages. According to \citet{conneau-etal-2020-unsupervised}, their model XLM-R outperforms previous models on a wide range of tasks, demonstrating that leveraging large-scale multilingual data can lead to improvements in cross-lingual understanding. 


Despite these advancements, significant challenges remain in achieving truly equitable performance across all languages and handling culturally specific contexts accurately \cite{zhang-etal-2020-improving}. 
Languages with limited digital text data still pose a considerable challenge for LLMs, often resulting in lower performance and less reliable outputs. Addressing this issue requires more inclusive data collection practices and further research into transfer learning techniques that can better utilise limited resources \cite{dogruoz-sitaram-2022-language}. Additionally, capturing cultural nuances and context-specific meanings is a complex task, as language is deeply intertwined with cultural and societal norms. Efforts to improve these aspects include developing more sophisticated algorithms and incorporating diverse and representative datasets \cite{dogruoz-etal-2023-representativeness}, ensuring that the benefits of multilingual LLMs are accessible to a broader range of users globally.


More recently, a plethora of social network datasets targeting mental health, have been available \cite{raihan2024mentalhelp}.
The authors gathered social media posts from Reddit and Twitter regarding depression, PTSD, schizophrenia, and eating disorders. Moreover, multiple models were fine-tuned on small-sized publicly available annotated mental health datasets by the authors to use them for labelling their introduced MentalHelp dataset. Nevertheless, the dataset includes only posts in English, and thus its use is restrictive, disallowing further research for multilingual scenarios.

\section{Proposed Methodology}

Our methodology leverages LLMs, in order to translate English social media posts to another language (Greek), and then to predict mental health conditions accordingly. Specifically, we translate the social media posts to Greek via an LLM, and we feed the resulting translations to a prompt that asks the LLM to predict specific severity levels of mental health conditions. We assess the LLM by comparing the predicted classes in both languages against the ground truth labels. We note that although our study focuses on Greek, our method is applicable to other language pairs as well. An illustration of the proposed approach for evaluating LLMs for multilingual detection of mental health conditions is shown in Figure~\ref{proposed}.



\begin{table*}[ht]\centering
  \resizebox{.8\textwidth}{!}{%
  \centering
  \begin{tabular}{lcccll}
    \hline
    \textbf{Dataset} & \textbf{Category} & \textbf{\#Classes} & \textbf{\#Instances} & \textbf{Labels (\#Support)} & \textbf{Prompt}\\
    \hline
    & \multirow{4}{*}{Depression} & \multirow{4}{*}{4} &  \multirow{4}{*}{3553} & Minimum (2587) & \multirow{4}{*}{\parbox{5cm}{``Categorise the  following text with 1 of the 4 depression severity levels (0: Minimum, 1: Mild, 2: Moderate, 3: Severe)"}} \\
    \textsc{DepSeverity} & & & & Mild (290) \\
    \citet{naseem2022early} & & & & Moderate (394) \\
    & & & & Severe (282) \\
    \hline
  \end{tabular}
  }
  \caption{The benchmark dataset used in our study along with statistics.}
  \label{tab:datasets}
\end{table*}

\section{Experiments}
We select the \textsc{DepSeverity} dataset of \citet{naseem2022early}, which consists of posts from the social media platform Reddit, regarding different levels of depression.
The posts (in English) are already labelled in terms of four levels of severity: minimal, mild, moderate, and severe depression. The majority of posts belong to the minimal severity level (Table~\ref{tab:datasets}) making it a highly imbalanced dataset.
We specifically selected this multi-class dataset to make the task more challenging for the LLM, as binary problems would have been easier to answer. 

We use GPT3.5-turbo \cite{brown2020} through its API to translate the posts and predict the labels. The temperature parameter is set to 0, so the outcome is reproducible, regarding translations and predictions. 
We approach the task with text classification, comparing the predicted classes with the ground-truth ones, reporting Precision, Recall and F1. 
We experiment with English as the source and Greek as the target language. 
The prompt we used to predict the severity levels is shown in Table~\ref{tab:datasets}.


\begin{table}[t]
\resizebox{\columnwidth}{!}{
    \centering
    \begin{tabular}{lccccccccc}
        \toprule
        & \multicolumn{3}{c}{\textbf{English}} & \multicolumn{3}{c}{\textbf{Greek}} \\
        \cmidrule(r){2-4} \cmidrule(r){5-7} 
        \textbf{Class} & \textbf{Pr} & \textbf{Rec} & \textbf{F1} & \textbf{Pr} & \textbf{Rec} & \textbf{F1} \\
        \midrule
        \sc minimum & 0.98 & 0.14 & \bf 0.25 & 0.99 & 0.07 & 0.14 \\
        \sc mild      & 0.04 & 0.15 & \bf 0.07 & 0.04 & 0.17 & 0.06 \\
        \sc moderate  & 0.13 & 0.22 & 0.17 & 0.14 & 0.55 & \bf 0.23 \\
        \sc severe    & 0.13 & 0.71 & \bf 0.22 & 0.16 & 0.28 & 0.20 \\
        \midrule
        \textbf{Macro avg} & 0.32 & 0.30 & \bf 0.17 & 0.33 & 0.27 & 0.16 \\
        \bottomrule
    \end{tabular}
    }
    \caption{\textbf{GPT-3.5 with 0-shot learning on \textsc{Dep-Severity}}, measuring Precision, Recall, and F1 per class in English and Greek. The last row shows the macro averages. The best F1 per class is shown in bold.}
    \label{results-depression}
\end{table}

\paragraph{Preliminary Prompting} Before exposing our LLM to any posts, definitions, or instructions, either for the translation or the classification task, we asked how it would classify posts to different levels of depression severity. 
The response of LLM was that it would initially try to identify language patterns associated with depression, such as:
\begin{itemize}
\setlength\itemsep{0.01cm}
\item Persistent negative emotions, such as sadness, or hopelessness.
\item Self-criticism or feelings of worthlessness.
\item Expressions of loneliness or social withdrawal.
\item Changes in behavior or routines, as in sleep patterns or appetite.
\item References to emotional pain or distress.
\end{itemize}
More specifically, it would try to adapt the four depression severity levels to fit the context of social media posts, as follows. 
\begin{itemize}
\setlength\itemsep{0.01cm}
\item Level 1 (Minimum): Posts with minimal or occasional expressions of sadness.
\item Level 2 (Mild): Posts indicating frequent negative emotions or noticeable changes in behavior.
\item Level 3 (Moderate): Posts suggesting significant impairment in daily functioning or clear signs of distress.
\item Level 4 (Severe): Posts indicating severe emotional distress, potential risk factors for self-harm, or complete social withdrawal.
\end{itemize}
We can infer that the LLM expects posts with very generic indications of negative signs.

\paragraph{Classification in the source language}\label{sssec:results_source}
Initially, we experimented with the data in their source language (English), to set the baseline performance. That is, no translation step has been performed at this stage. As we observe in Table~\ref{results-depression}, the best F1 is achieved for the lowest severity/indication (F1=0.25) and the next best for the highest severity (F1=0.22). The overall low performance (F1=0.17) can be attributed to the difficulty of the task of detecting specific levels of depression, which are considered less distinct compared to other conditions.
Therefore, it is likely more challenging for an LLM to distinguish these levels in user posts.

\begin{figure*}[t]
  \centering
  \includegraphics[width=.95\textwidth]{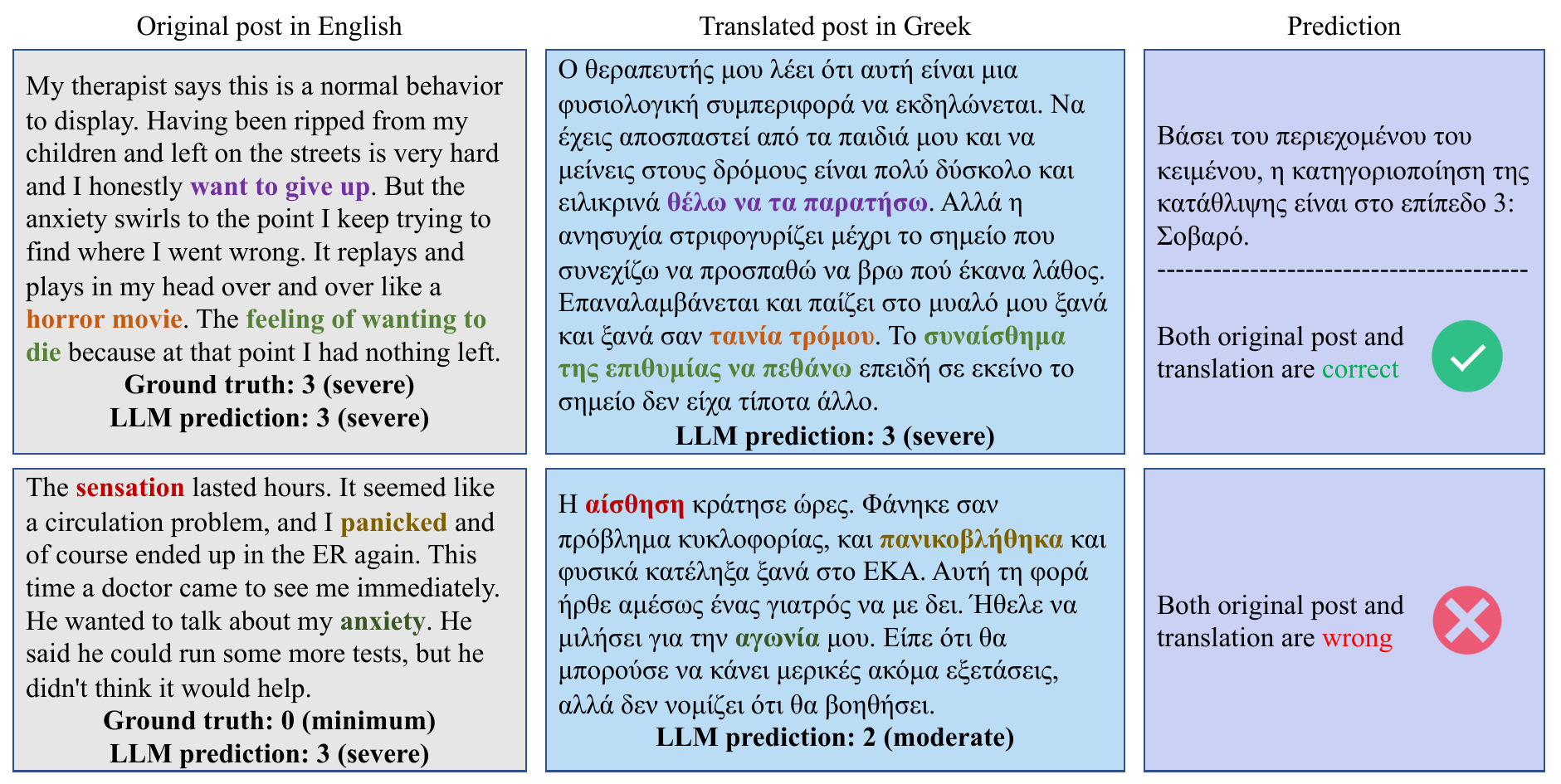}
  \caption{Example translation (from English to Greek), with similar colour used for original and translated words. 
  }
  \label{greek-depression-example}
\end{figure*}

\paragraph{Classification in the target language}
In Greek, the worst results are observed for the mild level (F1=0.06), similarly to English (F1=0.07). Overall, a drop in scores is observed across all classes except for the moderate level, where results improve (F1=0.23; from 0.17). Also, although the scores for the two edges remain relatively high, the score for the lowest severity dropped in Greek (F1=0.14).

\paragraph{Error Analysis}Mental health terminologies and nuances may not be well-represented in the available Greek corpora, making it difficult for an LLM to grasp the context accurately.
Figure \ref{greek-depression-example} presents two instances of the dataset and their corresponding translations in Greek. 
We marked words and their translations with similar colours for better visibility. 
Both translations appear to be accurate and convey the same meaning as the original Greek text. There are no significant differences that would alter the understanding of the texts.
The first segment contains explicit mentions of severe depression symptoms such as ``want to give up" and ``feeling of wanting to die.'' 
These statements clearly indicate a severe level of depression, which is why both the ground truth and prediction were classified as severe.
The second segment describes physical sensations, panic, and anxiety but does not express a severe depressive state. The ground truth classified this as minimal depression, likely because the primary issues are related to panic and anxiety rather than depression. 
The LLM predicted a moderate level of depression for the second segment, possibly because it picked up on the words ``panicked" and ``anxiety", which are associated with higher levels of distress. 
However, these symptoms are more indicative of anxiety disorders rather than depression. 
The discrepancy in the second prediction can be attributed to the LLM's interpretation of anxiety and panic as indicative of moderate depression, whereas the ground truth assessment considers these symptoms in the context of a panic or anxiety disorder with minimal depression. 

\paragraph{Cost of experiments}
The total cost of credits using the GPT3.5-turbo API was less than \$30 (US dollars), showing that minimal resources were required to conduct our experiments, without the need for expensive GPU infrastructure or fine-tuning. 
Our cost-saving methodology for utilizing resources efficiently is especially promising for extending medical data sets in English into other languages.

\section{Conclusion}
In our study, we focused on the ability of an LLM to predict the severity of depression in user-generated posts in English (source language) and in Greek (target language) when the posts are machine-translated by the same LLM. 
Our findings show that there is room for improvement in the source language (English) and that the edge classes are easier to handle. In the target language (Greek), results dropped for all but the moderate level, for which results increased considerably. 
Considering the varying performance of the LLM across the two languages, there is a need for utmost precautions not to rely on LLMs solely for translation in any healthcare setting.  
As stated by \newcite{stade2024large}, diagnosis of mental health should never be left alone to automatic systems, and it should never replace the diagnosis by human professionals, to avoid possible errors and/or misdiagnoses. 
Our approach, however, does not aim to assist the patients. By contrast, it is potentially useful to \emph{train} professionals in the mental healthcare domain, which can be vital for languages other than English. 

\section*{Acknowledgments}
In this project, John Pavlopoulos was partially supported by project MIS 5154714 of the National Recovery and Resilience Plan Greece 2.0 funded by the European Union under the NextGenerationEU Program.
We thank OpenAI for granting us free credits for research purposes.

\section*{Limitations}

\paragraph{Translation} In this work, we used a popular LLM like GPT3.5 to translate posts.
Translating using only an LLM and not having an expert or native-language human resources introduces a small loss of information that in some cases affects the final results.

\paragraph{Evaluation} Automatically evaluating the performance of LLMs is by definition a hard task.
In order to measure the performance we search for the label in the LLM output.
Whenever no label is detected we count it as the minimum label (class: 0) for the depression dataset and not suicidal (class: 0) for the suicide dataset.

\paragraph{Potential risks} The quality of publicly available datasets, especially in sensitive areas like the mental health care domain is of great importance for prediction tasks. The data sets we employed as a basis in our study, along with our created multilingual data should be used with utmost care and only for assisting the health care specialists instead of diagnosing patients directly.

\bibliography{acl_latex.bib}




\end{document}